\documentclass[11pt]{article}

\usepackage{acl}

\usepackage{times}
\usepackage{latexsym}

\usepackage[T1]{fontenc}

\usepackage[utf8]{inputenc}

\usepackage{microtype}

\usepackage{inconsolata}

\usepackage{graphicx}

\usepackage{booktabs}
\usepackage{amsmath}
\usepackage{epigraph}
\usepackage{subcaption}
\usepackage{longtable}
\usepackage[russian,english]{babel}
\usepackage{array}
\usepackage{tabularx}

\title{Instructions for *ACL Proceedings}

\author{Sarah Griebel\\
  University of Illinois Urbana-Champaign / Address line 1 \\
  Affiliation / Address line 2 \\
  Affiliation / Address line 3 \\
  \texttt{sarahg8@illinois.edu} \\\And
  Ted Underwood\\
  Affiliation / Address line 1 \\
  Affiliation / Address line 2 \\
  Affiliation / Address line 3 \\
  \texttt{email@domain} \\}

\title{Fluency and Faithfulness in Human and Machine Literary Translation}

\author{
  Sarah Griebel\thanks{Corresponding author.}
  \and
  Ted Underwood \\
  School of Information Sciences \\
  University of Illinois Urbana--Champaign \\
  \texttt{sarahg8@illinois.edu} \quad
  \texttt{tunder@illinois.edu}
}

\begin{document}
\maketitle

\begin{abstract}
Literary translation requires balancing target-language fluency with faithfulness to the source. Recent large language models (LLMs) often produce fluent translations, but it remains unclear whether fluency corresponds to semantic preservation in literary text. We examine this relationship using 130{,}486 translated paragraphs from 106 novels in 16 source languages, including human, Google Translate, and TranslateGemma translations. Fluency is measured as original-likeness with a translationese classifier trained on paragraph part-of-speech n-grams, and faithfulness with the automatic translation evaluation metric COMET-KIWI. We control for paragraph length and find a consistent negative correlation between fluency and faithfulness. The pattern appears for both human and Google Translate, but is weaker and often non-significant for TranslateGemma. These results show that segment length matters for automatic evaluation and suggest a tradeoff between fluency and faithfulness in literary translation.\footnote{Code and analysis for the project is available at \url{https://github.com/griebels/fluency-faithfulness-mt}.}
\end{abstract}

\epigraph{\textit{“Either the translator leaves the writer in peace as much as possible, and moves the reader toward him; or he leaves the reader in peace and moves the writer toward him.”}}{Friedrich Schleiermacher, \textit{On the Different Methods of Translating} (1813)\\transl. Susan Bernofsky}

\section{Introduction}

Since Schleiermacher’s famous formulation that a translator must either "move the reader toward the writer" or "move the writer toward the reader", translation theory has often treated fluency and fidelity as opposing poles. Yet this tension has rarely been examined empirically at scale.

Literary translation involves reconciling competing objectives. Translators must produce text that reads thoughtfully, smoothly, and accurately in the target language while preserving the semantic and stylistic content of the original. Figurative language--often present in literary text--adds a unique challenge. For example, a translation that closely follows the source structure of an idiom or metaphor may preserve meaning but sound unnatural or "foreign", while a highly fluent translation may involve paraphrasing or restructuring that alters semantic content, glosses it, or attempts to "domesticate" certain elements of the source text in the target tongue. (Translation theorists use different terms to describe this process; for example, foreignization and domestication \cite{schleiermacher_different_2012, venuti_translators_1995} or deforming tendencies \cite{berman_translation_2012}.)

Recent large language models (LLMs) have demonstrated strong performance in literary machine translation \cite{zhang_how_2025, karpinska_large_2023, wang_evaluating_2025}. Prior work has primarily evaluated LLM literary translation quality using reference-based metrics or human evaluation, as it is more difficult to assess results from quality estimation (reference-free) evaluation metrics in this domain \cite{karpinska_large_2023}. Therefore, less is known about how fluency and quality estimation metrics relate to one another in literary machine translation.


In this work, we examine the relationship between fluency and faithfulness metrics at paragraph level across human and machine translations. We analyze a corpus of 130{,}486 translation paragraphs from 106 novels spanning 16 source languages, with human translations, Google Translate outputs, and translations generated using TranslateGemma \cite{finkelstein_translategemma_2026}, a recent open translation model based on the Gemma3 architecture \cite{team_gemma_2025}. 


We operationalize fluency as original-likeness, measured using a translationese classifier trained to distinguish original English paragraphs from translated English paragraphs. To isolate syntactic fluency independent of semantic content, we train the classifier on part-of-speech (POS) anonymized text. We operationalize faithfulness using COMET-KIWI scores.

Using these measures, we compute correlations between fluency and faithfulness separately for human translations, LLM translations, and a traditional machine translation system. We hypothesize that LLM translations may exhibit a tradeoff between fluency and semantic faithfulness, reflected in a negative correlation between original-likeness and adequacy scores, while human translations may exhibit weaker or different relationships.

Our contributions are as follows:

\begin{itemize}
\item We introduce a large-scale paragraph-level analysis of fluency and faithfulness in literary translation, spanning over 130{,}000 paragraphs across human, Google Translate, and LLM translations.
\item We propose a scalable framework for measuring fluency independent of semantic fidelity using POS-based literary translationese classification with book-level held-out evaluation.
\item We apply reference-free quality estimation (COMET-KIWI) to get a proxy for faithfulness across human, Google Translate, and LLM literary translations.
\item We provide empirical evidence that a negative relationship exists between fluency and faithfulness signals and is more distinctly observed among human literary translations of similar lengths, signaling that COMET-KIWI does not align with POS-based signals of fluency.
\end{itemize}

\section{Related Work}

\subsection{Fluency and faithfulness in machine translation evaluation}

Machine translation quality has traditionally been evaluated along two primary dimensions: fluency, referring to the naturalness and grammaticality of the target text, and faithfulness (or adequacy), referring to preservation of source meaning. Human evaluation protocols commonly distinguish these dimensions explicitly, recognizing that highly fluent translations may diverge semantically from the source, while highly faithful translations may exhibit syntactic or stylistic interference from the source language.

\subsection{Translationese and original-versus-translated classification}

Translated text exhibits systematic differences from text written originally in a target language, a phenomenon known as translationese. Prior work has shown that translated text can be distinguished from original text using lexical, syntactic, and distributional features, including function word frequencies \cite{koppel_translationese_2011}, string kernel methods \cite{popescu_studying_2011}, part-of-speech perplexity \cite{bizzoni_how_2020}, mean word length, syllable ratio, character n-grams, and others \cite{volansky_features_2015}. These differences typically arise from structural interference from the source language, simplification, and normalization during translation.


In this work, we build on translationese detection methodology to operationalize fluency as original-syntactic-likeness, measured by a classifier trained to distinguish original English paragraphs from translated English paragraphs using features built on ordered parts-of-speech input. We do not aim to produce a perfectly accurate classifier; rather, our goal is to produce a highly interpretable one.

\subsection{Faithfulness measures in literary machine translation}

Automatic translation evaluation metrics such as BLEU \cite{papineni_bleu_2002} and METEOR \cite{banerjee_meteor_2005} are limited, as they rely on reference comparisons and often fail to produce high overlap with human judgment. More recent neural metrics such as COMET \cite{rei_comet_2020} address these limitations by predicting translation quality using multilingual encoders. COMET fine-tunes pretrained multilingual language models to estimate translation adequacy based on reference and target text, and has been shown to have high overlap with human evaluation \cite{lee_survey_2023}. Its reference-free variant, COMET-KIWI \cite{rei_cometkiwi_2022}, enables direct estimation of semantic faithfulness without requiring gold reference translations. 



While neural evaluation metrics can be considered "black box" as to what their outputs are actually measuring \cite{karpinska_demetr_2022}, COMET outputs were demonstrated to rely heavily on multilingual sentence representations derived from pretrained encoders \cite{rei_inside_2023}. Therefore, COMET provides a scalable proxy for semantic adequacy, which we interpret here as a signal of faithfulness when analyzed alongside fluency measures.

Reference-based evaluation metrics work well to evaluate the performance of machine translation systems, but they impose a difficulty when comparing human to machine translations by requiring a gold-standard reference. Quality estimation metrics in machine translation are a valuable tool, as they do not require reference translations to compare source and target texts. Moreover, recent advances have shown overlap between QE results and human judgment in machine translation tasks \cite{zerva_findings_2024, specia_findings_2020}. By correlating COMET-KIWI with syntactic fluency, we are able to study how it performs for literary translation evaluation.

\section{Data}

\subsection{Aligned literary translation corpus}

We use the Par3 Dataset from \citet{thai_exploring_2022}, a paragraph-aligned corpus of literary translations spanning 113 novels and 122{,}819 source paragraphs. The corpus includes source texts in 16 languages: Czech, Dutch, French, German, Hungarian, Italian, Japanese, Norwegian, Persian, Polish, Portuguese, Russian, Spanish, Swedish, Tamil, and Chinese. For each source paragraph, the dataset includes two or more human translations (maximum 5), as well as machine-generated translations from Google Translate. 

We generate LLM translations using TranslateGemma, a fine-tuned translation model suite built on the architecture of Gemma3, evaluated on WMT benchmarks across 55 languages that demonstrates substantial improvements over the base Gemma model \cite{finkelstein_translategemma_2026}. We use the 4-billion-parameter model version to retain low overhead compute. We run inference on the model through Ollama, an open-source tool to run models locally. Translation prompts are given at the paragraph-level, using the recommended prompt technique for the model.

Paragraph alignment is provided as part of the dataset release and further verified using multilingual embedding similarity filtering (Section~\ref{sec:guardrails}).

Human translations in the corpus span a range of publication periods. Because translation publication dates are not made available with the dataset, we do not condition analyses on this. However, we do discuss implications of translation publication dates on fluency measures (Section~\ref{sec:limitations}).

\paragraph{Duplicate and overlapping works.}

The Par3 corpus contains multiple entries corresponding to the same underlying literary work, including cases where a single novel appears multiple times. To avoid overweighting individual works and introducing near-duplicate material into classifier training, we identify and remove entries that represent the same original text. 

When multiple entries corresponded to a single work (e.g., \textit{Les Misérables 1, Les Misérables 2}), we retain only the first representative volume. In contrast, distinct works by the same author (e.g., separate novels within a series) are retained. This filtering reduced the original dataset from 113 books to 106 books.

\subsection{Original English comparison corpus}

To train a translationese classifier, we use a corpus of 115 English-language novels (i.e., originally written in English) published between 1800 and 1930. In selecting the earliest dates for our original dataset, we aimed to gather texts that would align with an empirical translation lag from the source texts in Par3, which were published between 1399\footnote{Though the earliest source text in the Par3 dataset was published in 1399, the majority (nearly 90\%) were published after 1800.}--1982 but mostly clustered in the middle. Our 1930 ceiling reflects standard U.S. public domain constraints.


Paragraphs from this corpus serve as positive examples of original English, while translated paragraphs serve as negative examples. 

\section{Methods}

We measure fluency and faithfulness independently using translationese classification and quality estimation, respectively. This allows us to examine their relationship across human and machine literary translations.

\subsection{Fluency as original-likeness}

We operationalize fluency as \emph{original-likeness}, defined as the degree to which a translated paragraph resembles original English text. 


To isolate syntactic structure independent of semantic content, we anonymize each paragraph by replacing tokens with their fine-grained part-of-speech (POS) tags, using a spaCy model. For example:

\begin{quote}
Original: ``Ivan asked Dmitry to meet him at the inn for lunch today?'' Alyosha asked quickly.

POS-anonymized: ``NNP VBD NNP TO VB PRP IN DT NN IN NN NN .'' NNP VBD RB .
\end{quote}

We use the POS-based classifier as our fluency metric because it isolates syntactic structure independent of lexical content, ensuring that fluency reflects structural naturalness rather than topical or lexical differences.

\paragraph{Minimum paragraph length filtering.}

To ensure that the classifier captures meaningful linguistic structure, we exclude paragraphs shorter than a minimum 20 word threshold. Short segments such as chapter headings (e.g., ``IV'' or ``Chapter VII'') or ornamental separators do not contain sufficient syntactic structure to reliably distinguish original from translated text. Including such segments would introduce noise and potentially bias the classifier toward non-linguistic artifacts.

\paragraph{Classifier training, preprocessing, and cross validation.}

We train a logistic regression classifier using TF-IDF features. Features include unigram, bigram, and trigram n-grams, with a maximum vocabulary size of 20{,}000 features. The classifier uses L2 regularization with inverse regularization strength $C=10.0$, \texttt{class\_weight=balanced}, and a maximum of 2000 optimization iterations.

To minimize the influence of paragraph length on classifier predictions, we downsample the translated class using ten paragraph-length bins derived from the original-English text distribution. For each bin, we randomly sample an equal number of translated paragraphs to match the number of original paragraphs in that bin. This procedure reduces the dataset used for comparison from 264,487 to 130,486 paragraphs.

Because each source paragraph may have multiple translated versions (including human translations, Google Translate, and TranslateGemma outputs), the translated class contains varying numbers of observations derived from the same underlying content. To prevent paragraphs with more translation variants from disproportionately influencing training, we assign each translated paragraph a sample weight inversely proportional to the number of translations available for its source paragraph. Specifically, if a source paragraph has $n$ translated versions, each translation is assigned weight $1/n$. Original English paragraphs are assigned uniform weight.

To prevent leakage of book-specific stylistic features, we perform 10-fold group cross-validation using books as grouping units, ensuring that all paragraphs from a given book appear exclusively in either training or evaluation folds.

Because source languages are unevenly represented, we additionally stratify books by source language when constructing folds. Languages with only one or two books are combined into a single ``rare languages'' stratum. This ensures that each fold contains representation from as many languages as possible while preserving strict book-level separation, and prevents classifier performance from being driven by language-specific artifacts or imbalanced language representation.

Each paragraph receives an out-of-fold prediction corresponding to the probability of belonging to the translated class, $P(\text{translation} \mid x)$. We define fluency as:

\begin{equation}
\text{Fluency}(x) = 1 - P(\text{translation} \mid x)
\end{equation}

Higher scores indicate greater similarity to original English and therefore higher fluency.

\paragraph{Translationese classifier performance}

Table~\ref{tab:classifier_performance} summarizes the performance of the translationese classifier. The classifier achieves strong performance, confirming that translated paragraphs exhibit systematic structural differences from original English at paragraph level. 

\begin{table}[h]
\centering
\small
\begin{tabular}{lccc}
\toprule
Features & Accuracy & Macro F1 & AUC \\
\midrule
POS abstraction & 0.760 & 0.753 & 0.847 \\
\bottomrule
\end{tabular}
\caption{Performance of translationese classifier under book-level cross-validation.}
\label{tab:classifier_performance}
\end{table}

\subsection{Faithfulness as semantic adequacy}

We measure faithfulness using COMET-KIWI, a neural metric trained to predict translation quality without reference translations. COMET-KIWI takes as input the source paragraph and translated paragraph and produces a scalar adequacy score between 0-1.


We use the \texttt{Unbabel/wmt22-cometkiwi-da} model for quality estimation.

\paragraph{Sanity check: COMET-KIWI and roundtrip semantic consistency.}

To verify that COMET-KIWI provides a reliable signal of semantic faithfulness in our dataset, we compare COMET-KIWI scores with roundtrip semantic similarity. For each TranslateGemma output, we generate a roundtrip translation by translating the English translation output back into the source language. We compute cosine similarity between the original source paragraph and the roundtrip text using BGE-M3 multilingual sentence embeddings \cite{chen_m3-embedding_2024}, which has shown strong performance for a range of multilingual tasks and supports long inputs.

\begin{figure}
\centering
\includegraphics[width=\columnwidth]{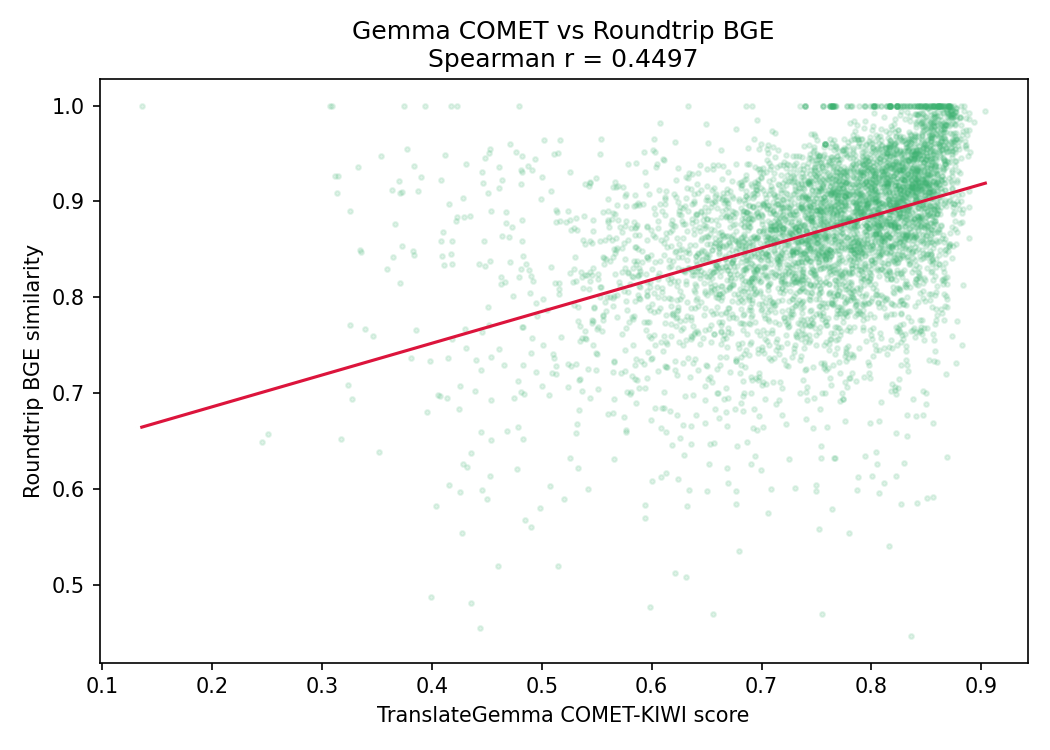}
\caption{Correlation between COMET-KIWI scores and roundtrip semantic similarity for TranslateGemma translations. Higher COMET-KIWI scores correspond to greater semantic consistency under roundtrip translation.}
\label{fig:comet_roundtrip_corr}
\end{figure}

We observe a strong positive correlation between COMET-KIWI scores and roundtrip semantic similarity (Figure~\ref{fig:comet_roundtrip_corr}), indicating that COMET-KIWI captures semantic preservation consistent with independent embedding-based similarity measures. This provides additional support for using COMET-KIWI as a reference-free proxy for semantic faithfulness in our literary translation task.

\subsection{Guardrail filtering for alignment and translation validity}
\label{sec:guardrails}

To ensure reliable comparison, we apply several filtering criteria to remove paragraphs with potential misalignment or translation errors.

To make sure LLM translations are valid, we apply a length consistency filter, excluding paragraphs where the TranslateGemma output exceeds the Google Translate output by more than 500 characters, which may indicate degenerate generation. This consistency filter removes 42 LLM translations.

We then employ paragraph alignment validation, filtering sentences below a cosine similarity threshold, as described in detail below. 

\paragraph{Alignment validation and filtering.}

Because paragraph alignment in literary corpora may occasionally contain errors or mismatches, we use embedding similarity to identify potentially misaligned translations. We compute cosine similarity between source and translated paragraphs using BGE-M3 embeddings for all translation types.

\begin{figure}
\centering
\includegraphics[width=\columnwidth]{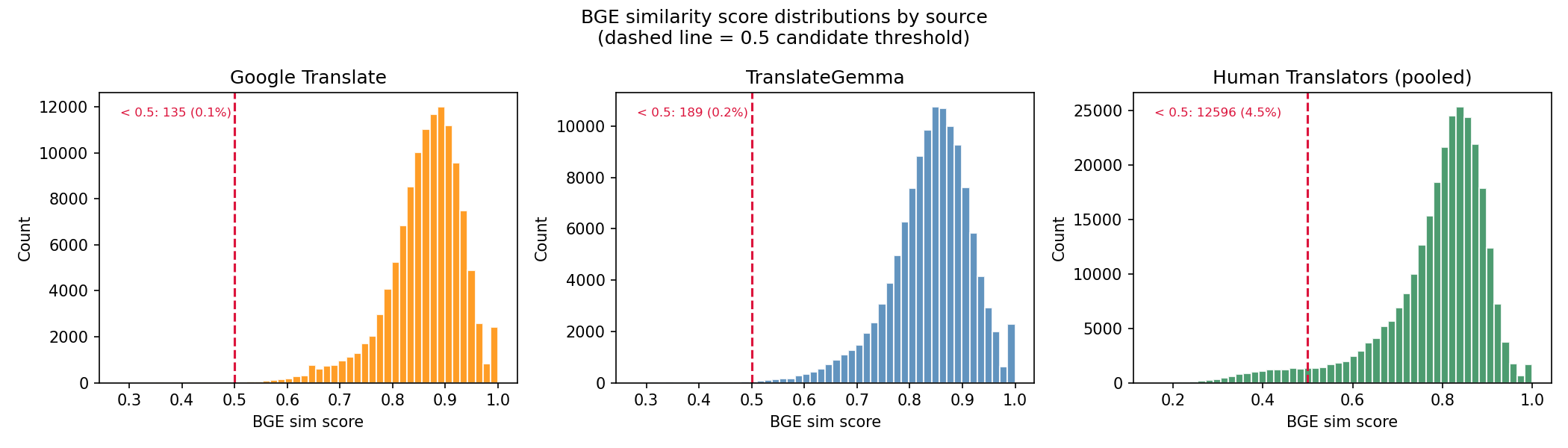}
\caption{Distribution of source–translation cosine similarity using BGE-M3 embeddings. Low-similarity tail primarily reflects misaligned or invalid paragraph pairs.}
\label{fig:bge_distribution}
\end{figure}

The distribution of similarity scores reveals a long tail of low-similarity examples, particularly among human translations (Figure~\ref{fig:bge_distribution}). To assess whether these low-similarity cases reflect true semantic divergence or alignment errors, we manually inspect a random sample of 30 paragraphs from the low-similarity tail. We find that over 93\% of these cases exhibit clear alignment errors, primarily unrelated content. Table~\ref{tab:misalignment_examples} shows representative examples of these misalignments.

\begin{table*}[t]
\small
\setlength{\tabcolsep}{4pt}
\renewcommand{\arraystretch}{1.15}
\centering

\begin{tabularx}{\textwidth}{
>{\raggedright\arraybackslash}p{0.26\textwidth}
>{\raggedright\arraybackslash}p{0.27\textwidth}
>{\raggedright\arraybackslash}p{0.27\textwidth}
>{\centering\arraybackslash}p{0.15\textwidth}
}

\toprule
\textbf{source\_para} &
\textbf{gt\_para} &
\textbf{translation} &
\textbf{cosine similarity} \\
\midrule

* * * * * &
* * * * * &
Go ask the artillerymen of Saint-Roch \dots &
0.422 \\
\midrule

68. Par exemple la Sphynge, une lionne ail\'ee \`a t\^ete de femme, qui interroge \OE dipe. &
68. For example the Sphynge, a winged lioness with the head of a woman, who questions Oedipus. &
CHAPTER XIV &
0.427 \\
\midrule

\foreignlanguage{russian}{— Кому же они привезли?} &
Whom did they bring? &
``You have heard it all. &
0.409 \\

\bottomrule
\end{tabularx}
\caption{Examples of low-similarity paragraph pairs manually identified as misaligned. In these cases, the "translation" field contains content inconsistent with the aligned source paragraph (e.g., chapter headings or unrelated narrative content).}
\label{tab:misalignment_examples}
\end{table*}

Based on this analysis, we apply a similarity threshold to exclude paragraphs with low embedding similarity, ensuring that subsequent analyses focus on reliably aligned translations. Importantly, we use embedding similarity rather than COMET-KIWI scores for filtering to avoid circularity, as COMET-KIWI serves as our primary faithfulness metric.

Overall, filtering on this cosine similarity threshold reduced the translation corpus by 2.0\%. This primarily affected human translations, rather than translations from Google Translate or TranslateGemma.

\section{Results}

\subsection{Raw fluency and faithfulness scores by paragraph and by book}

\begin{figure*}[t]
    \centering
    \includegraphics[width=\textwidth]{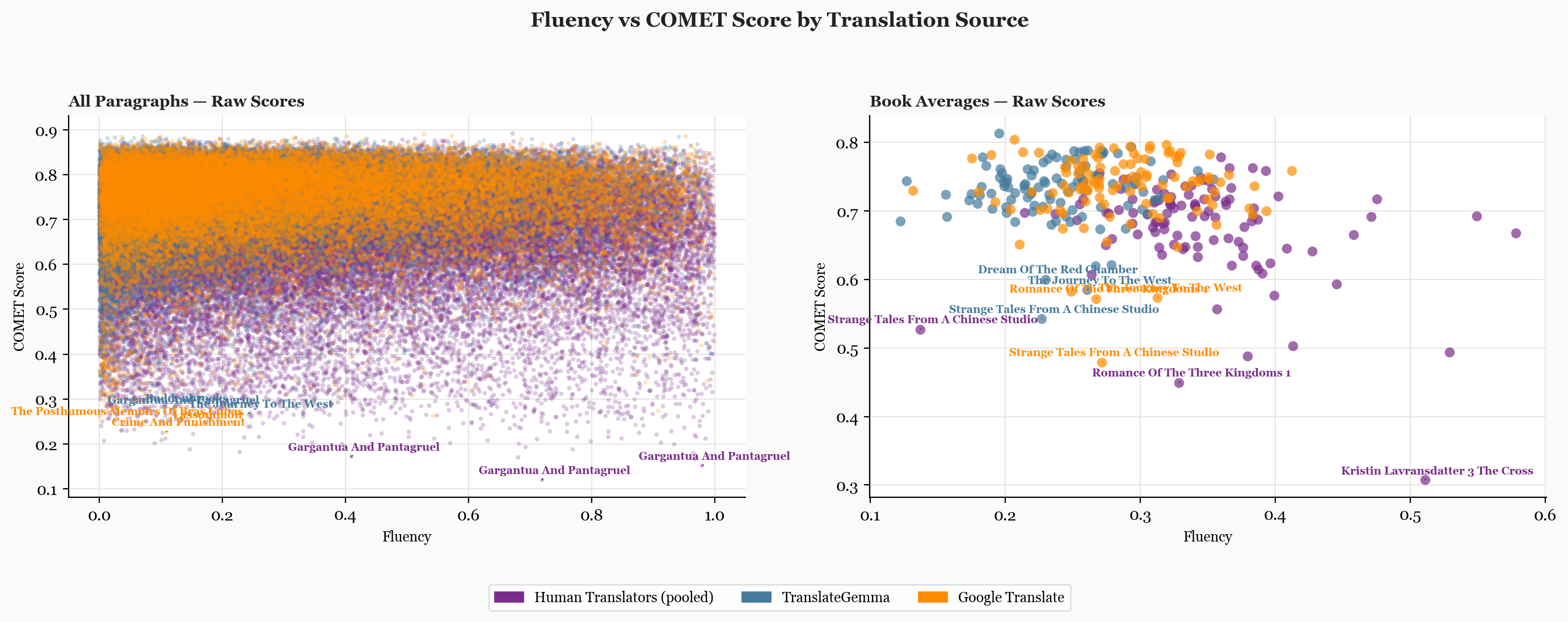}
    \caption{Fluency vs.\ COMET-KIWI score by translation source. 
    Points represent individual paragraphs (left) and books (right), colored by source: human translators (purple), 
    TranslateGemma (blue), and Google Translate (orange). For human translations, scores are averaged across multiple translations.
    Labeled points indicate the largest outliers by residual distance from the per-source fit line.}
    \label{fig:raw_scores}
\end{figure*}

In Figure~\ref{fig:raw_scores} we observe that human translations appear at both higher and lower faithfulness thresholds of the graph, while the Google Translate and TranslateGemma results cluster near the top. The contrast becomes more stark when reviewing the bottom right quadrant, indicating lower faithfulness but high fluency. In fact, 27.53\% of human translations in our dataset score above 0.5 in our fluency measure, while only 21.37\% and 15.26\% of Google Translate and TranslateGemma (respectively) score above this threshold.



\subsection{Fluency--faithfulness association is length-conditioned}

We examine the relationship between fluency and faithfulness using Spearman correlations between fluency (1 - $P(\text{translation} \mid x)$) and COMET-KIWI adequacy scores. 

When analyzing all paragraphs, we observe a small overall association ($\rho = -0.0374$). However, stratifying by paragraph length reveals a clear difference between long and short paragraphs: For paragraphs shorter than 100 words, the correlation between fluency and COMET-KIWI tends to be larger (Spearman $\rho = -0.086$, $p < 0.001$). For paragraphs of 100 words or longer, the relationship reduces to nearly zero (Spearman $\rho = -0.0151$, $p < 0.05$). 

\subsection{Paragraph length correlates with both metrics}

The length-stratified difference suggests that paragraph length may confound the fluency--faithfulness relationship. We therefore examine correlations between paragraph length and each metric across the full dataset.

Indeed, we find that paragraph length is negatively correlated with COMET-KIWI scores (Spearman $\rho = -0.2641$, $p < 0.001$) and with fluency (Spearman $\rho = -0.1565$, $p < 0.001$). Notably, paragraph length is essentially uncorrelated with embedding-based alignment similarity used in our guardrail filtering (Spearman $\rho = 0.0074$), suggesting that the negative relationship between length and COMET-KIWI is not driven by increasing misalignment in longer segments.

Together, these results indicate that paragraph length exerts substantial influence on both fluency and adequacy metrics.

\subsection{Controlling for length increases the association between fluency and faithfulness}

To estimate the direct relationship between fluency and faithfulness independent of paragraph length, we compute a partial Spearman correlation controlling for length. The association between fluency and COMET-KIWI strengthens overall (partial Spearman $\rho = -0.0827$, $p < 0.001$). Parceling this out by translation source, the signal is weakly positive for TranslateGemma ($\rho = 0.0129$) $p < 0.05$). It is negative for human translations ($\rho = -0.0783$ $p < 0.001$) and Google Translate ($\rho = -0.0512$ $p < 0.001$). 

\begin{figure}[t]
\centering
\includegraphics[width=0.8\linewidth]{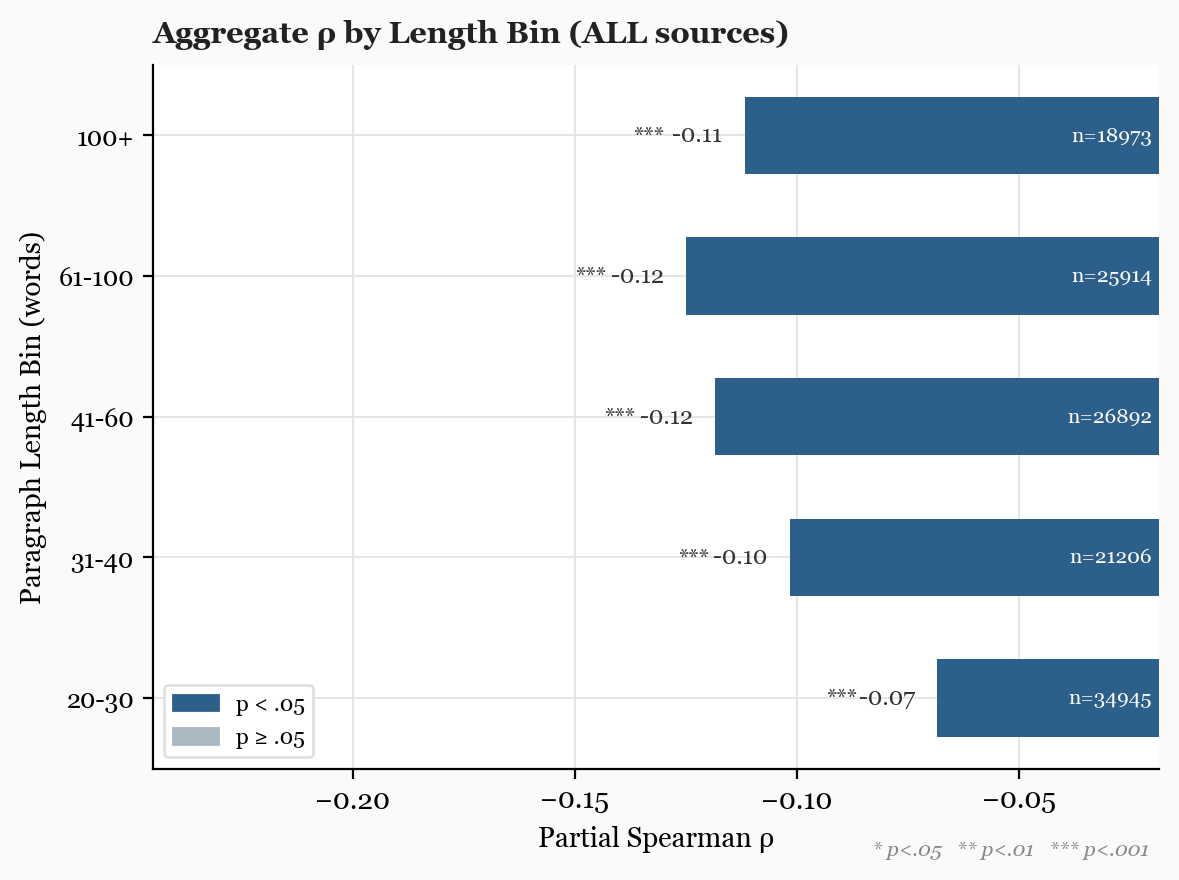}
\caption{Partial Spearman correlation between fluency and COMET-KIWI after controlling for paragraph length, computed separately for paragraph length bins. The correlation is modest but consistently negative across bins, with the strongest association appearing for mid-length segments (61--100 words).}
\label{fig:length_bins}
\end{figure}

Stratifying the analysis by paragraph length reveals a similar pattern. Within each length bin, the partial correlation between fluency and COMET-KIWI remains negative, ranging from $\rho = -0.068$ for shorter segments (20--30 words) to $\rho = -0.125$ for mid-length segments (61--100 words), before slightly declining in association for the longest segments (100+ words). Figure~\ref{fig:length_bins} illustrates these bin-level correlations.

\begin{figure}[t]
\centering
\includegraphics[width=0.9\linewidth]{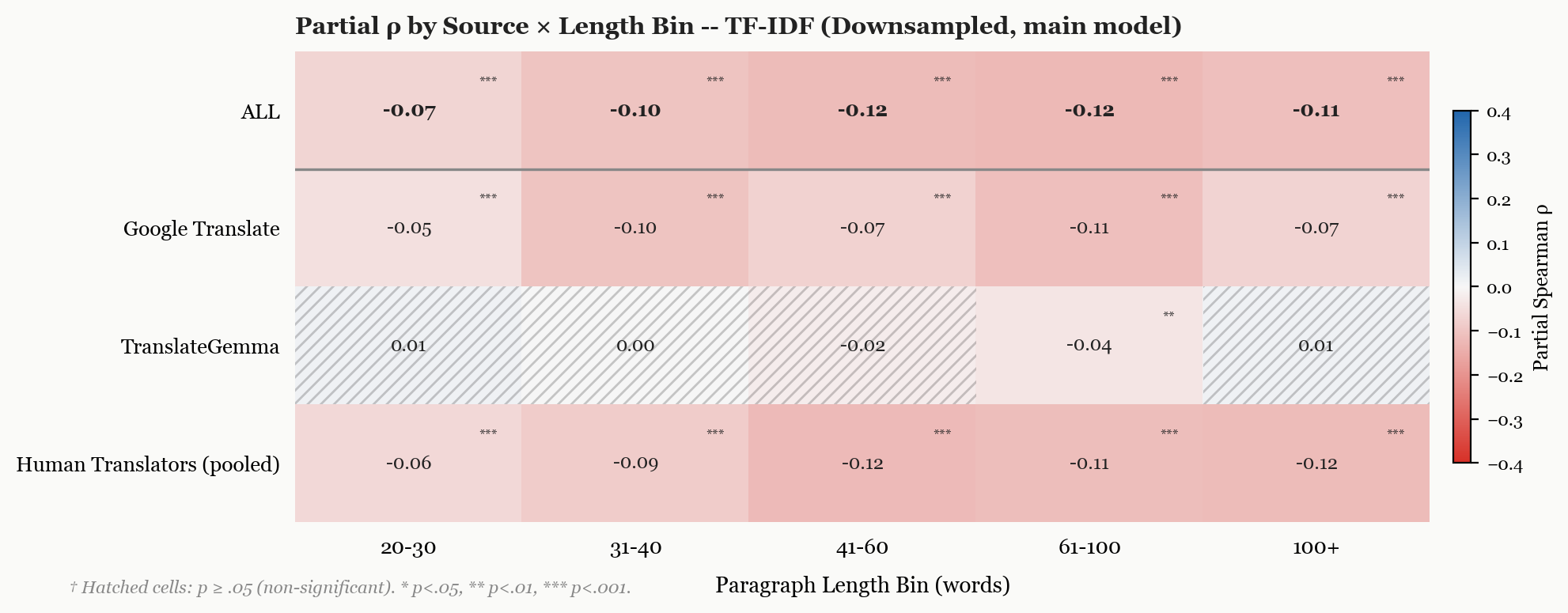}
\caption{Partial Spearman correlations between fluency and COMET-KIWI by paragraph source and length bin. Stars indicate statistically significant correlations ($p < 0.05$). Human and Google Translate translations show a broadly consistent negative association across bins, while the LLM-generated translations exhibit weaker and often non-significant relationships.}
\label{fig:source_heatmap}
\end{figure}

Interestingly, the pattern within length bins is also similar to the global pattern: TranslateGemma tends to have a weaker, sometimes positive correlation, while human translations show stronger negative correlations. More fine-grained variation across translation sources is shown in Figure~\ref{fig:source_heatmap}.

These patterns suggest that the global correlation is affected by the differences between human and machine translation outputs for these literary paragraphs: while human translations show a consistent negative relationship, both TranslateGemma and Google Translate tend to show a weaker, sometimes positive relationship between fluency and faithfulness. 



\section{Conclusion}

This study examined the relationship between syntactic fluency and semantic faithfulness in literary translation using a large corpus of paragraph-aligned texts translated by humans, Google Translate, and TranslateGemma. 

Across the full dataset, COMET-KIWI scores exhibited a negative correlation with paragraph length, indicating that longer segments tend to receive lower predicted adequacy scores. To control for this confounding factor, we regressed both fluency and COMET-KIWI on paragraph length and analyzed the residual signals. The resulting length-controlled measures reveal a modest but consistent negative association between syntactic fluency and semantic faithfulness.

Additional analyses further support this pattern. When paragraphs are grouped into length bins, the negative relationship between fluency and COMET-KIWI becomes more pronounced across most bins. This effect is visible for both human and machine translations, though the signal remains weaker and statistically non-significant for TranslateGemma.

Taken together, these findings suggest a small but systematic tension between producing syntactically native-like English and preserving semantic correspondence with the source text. While the magnitude of the relationship is modest, the consistency of the pattern across analytical approaches indicates that the two signals are not entirely independent.

This pattern also has implications for the interpretation of automatic evaluation metrics. Because COMET-KIWI is trained to approximate adequacy judgments, it may implicitly favor translations that preserve structural or lexical patterns closely aligned with the source text. As a result, the metric may partially reward signals associated with translationese when those signals correlate with surface-level source fidelity.

\section{Discussion}

One possible interpretation of these findings is that COMET-KIWI captures aspects of translation quality that are not aligned with the notion of fluency operationalized in this study. Our fluency measure reflects the degree to which a paragraph's syntactic structure resembles that of original English prose. By contrast, COMET-KIWI is trained to estimate semantic adequacy relative to the source text. The weak negative association observed here therefore suggests that producing highly native-like syntactic patterns does not necessarily coincide with maximizing semantic correspondence as predicted by COMET-KIWI.

The behavior of the fluency classifier also provides a lens to examine stylistic differences between human and machine translations. Because the model was trained on original English texts published between 1800 and 1930, it captures the syntactic distributions characteristic of that period. Although exact publication dates are unavailable for the human translations in Par3, nearly 90\% of the source texts were published after 1800, suggesting that at least some translations may have been produced within a similar historical context. In addition, translators working on historical literary texts may intentionally approximate earlier stylistic conventions in the target language. Contemporary machine translation systems and large language models, by contrast, are typically optimized for modern standard English and may only mirror the language used in popular translations --- that is, those that appear most frequently in their training data \cite{arnold_echoes_2025}. It is unclear whether modern models would naturally reproduce historical stylistic patterns absent the high-quality human examples in their training data. Consequently, for books with fewer human examples, model outputs may appear less "original-like" to the classifier, even when they are grammatically fluent.


More broadly, these findings echo long-standing discussions in translation theory regarding the tension between producing fluent target-language prose and preserving the foreign structure of the source text. While the correlations observed here are modest, they suggest that computational metrics may capture traces of this trade-off in large-scale translation data.

\section*{Limitations}
\label{sec:limitations}


Some limitations should be considered when interpreting these results.


First, the metadata available for the dataset is limited. Information about translators and publication dates of individual translations is not consistently available. As a result, we are unable to analyze how translator identity, translation era, or editorial practices might influence the observed relationship between fluency and faithfulness.

Second, the fluency measure used in this study captures only syntactic signals derived from part-of-speech patterns. While this abstraction helps isolate structural aspects of translationese, it does not account for lexical choices, stylistic register, or rhetorical effects that may also influence perceptions of fluency. Future work could incorporate additional stylistic features or human evaluation to further contextualize these findings.


\bibliography{ACL_NLP4DH_Translationese}
\clearpage
\appendix

\section{Classifier Variants}
\label{app:classifier_variants}

To evaluate the robustness of our findings, we train several alternative versions of the translationese classifier. These experiments test whether the observed relationship between fluency and paragraph length depends on specific modeling choices, particularly the feature representation and the use of downsampling.

Our primary classifier uses TF–IDF weighting on POS n-gram features. One concern with this representation is that TF–IDF emphasizes rare feature combinations, which could cause the classifier to rely disproportionately on infrequent POS patterns rather than capturing broader syntactic tendencies associated with translated text. To examine this possibility, we train additional classifiers using raw frequency counts of POS n-grams instead of TF–IDF weighting.

We also test whether our downsampling strategy, which balances paragraph length distributions across classes, influences the relationship between fluency and paragraph length. To assess this, we train models both with and without downsampling.

Specifically, we train Logistic Regression classifiers under the following configurations:

TF–IDF, full dataset: all samples (264,487 paragraphs), without downsampling
Count features, downsampled dataset: POS frequency counts with length-balanced downsampling
Count features, full dataset: POS frequency counts using all samples

Across these variants, we observe that the correlation between paragraph length and predicted fluency increases relative to the primary model, indicating that length sensitivity is not unique to the TF–IDF representation.

Table~\ref{tab:classifier_variants} summarizes the results.

\begin{table}[b]
\centering
\small
\setlength{\tabcolsep}{4pt}
\caption{Robustness checks for alternative translationese classifiers. Length--Fluency $\rho$ denotes the Spearman correlation between paragraph length and predicted fluency. Partial $\rho$ denotes the partial Spearman correlation between fluency and COMET-KIWI controlling for paragraph length.}
\label{tab:classifier_variants}

\begin{tabular}{llccc}
\hline
Features & Sampling & Samples & Length--Fluency $\rho$ & Partial $\rho$ \\
\hline
TF--IDF & Downsampled & 130486 & -0.1565 & -0.0827 \\
TF--IDF & Full & 264487 & -0.2470 & -0.0344 \\
Count & Downsampled & 130486 & -0.3631 & -0.0719 \\
Count & Full & 264487 & -0.4280 & -0.0444 \\
\hline
\end{tabular}
\end{table}

Despite these differences, the core findings remain stable across classifier variants. In particular, the negative partial Spearman correlation between fluency and COMET-KIWI persists, and the TranslateGemma condition remains largely non-significant across paragraph-length bins.

\begin{figure*}[t]
\centering

\begin{subfigure}{0.48\textwidth}
    \centering
    \includegraphics[width=\linewidth]{panel_c_heatmap_pooledHT_downsamp.png}
    \caption{TFIDF (downsampled)}
\end{subfigure}
\hfill
\begin{subfigure}{0.48\textwidth}
    \centering
    \includegraphics[width=\linewidth]{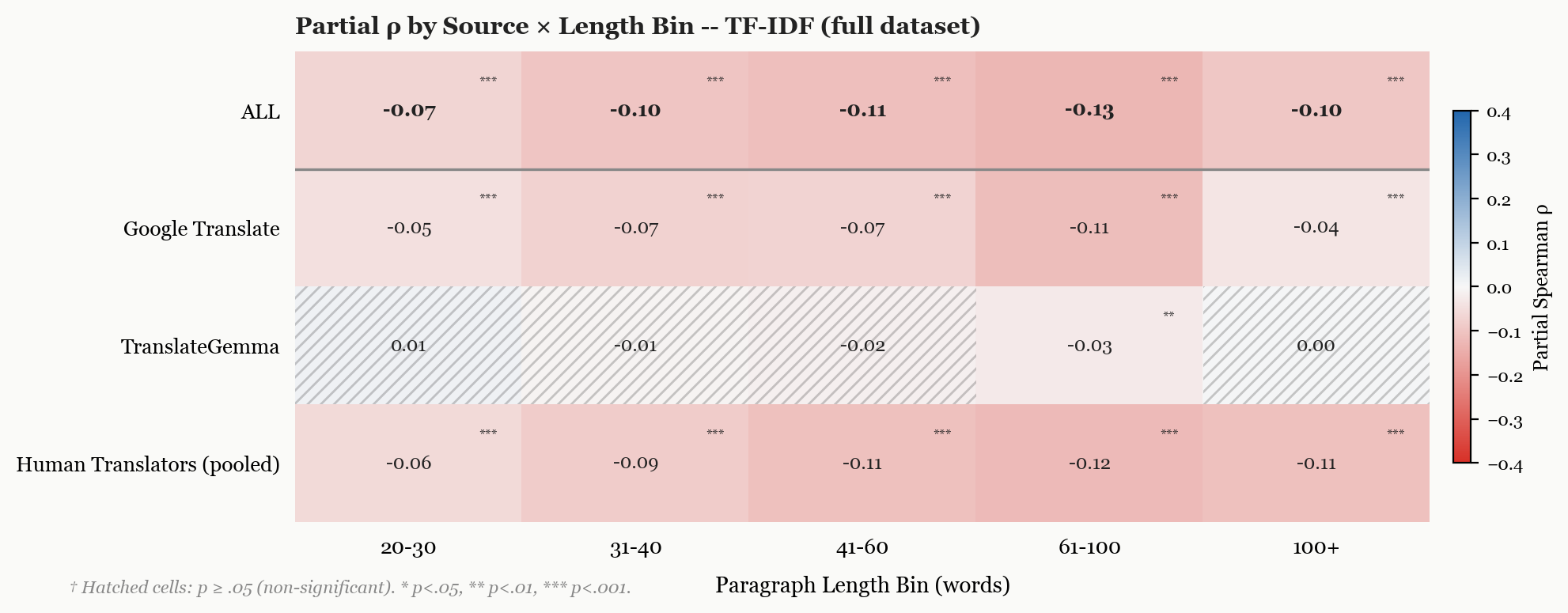}
    \caption{TFIDF (full dataset)}
\end{subfigure}

\vspace{0.5em}

\begin{subfigure}{0.48\textwidth}
    \centering
    \includegraphics[width=\linewidth]{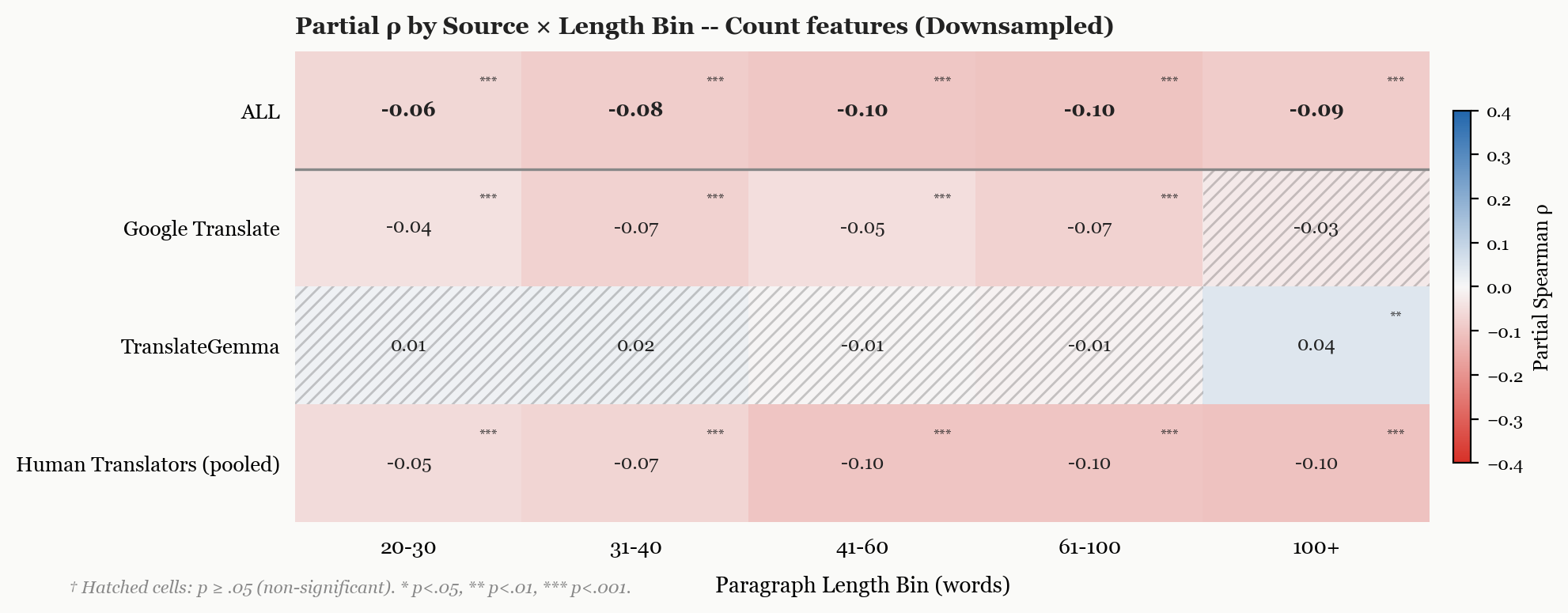}
    \caption{Count features (downsampled)}
\end{subfigure}
\hfill
\begin{subfigure}{0.48\textwidth}
    \centering
    \includegraphics[width=\linewidth]{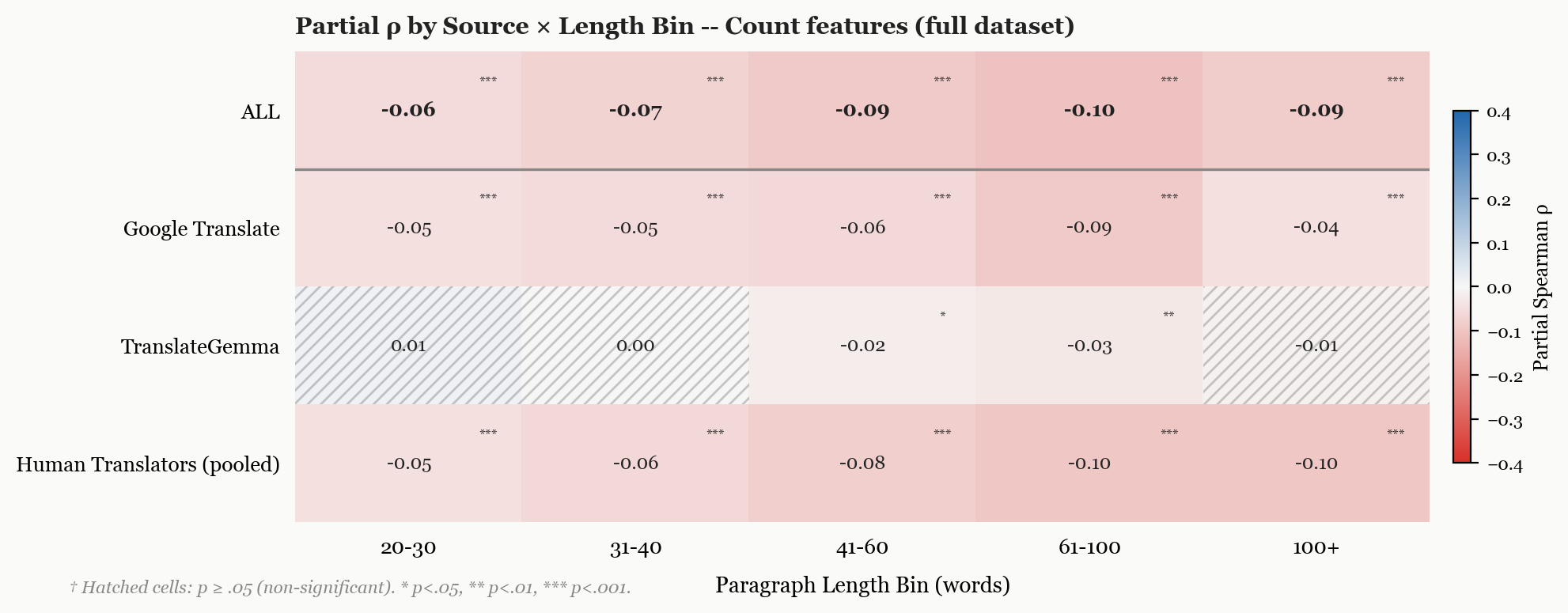}
    \caption{Count features (full dataset)}
\end{subfigure}

\caption{Heatmaps showing partial Spearman correlations between fluency and COMET-KIWI across paragraph-length bins for different classifier variants. The overall pattern across paragraph-length bins, as well as the non-significant results for TranslateGemma, remains consistent across feature representations and sampling strategies.}
\label{fig:robustness_heatmaps}

\end{figure*}

These results are visualized in Figure~\ref{fig:robustness_heatmaps}.

\end{document}